\documentclass[letterpaper, 10 pt, conference]{ieeeconf}  
\IEEEoverridecommandlockouts                              
\usepackage{graphics}    
\usepackage{times}       
\usepackage{amsmath}     
\usepackage{amssymb}     
\usepackage{graphicx}
\usepackage{algorithm}
\usepackage[noend]{algpseudocode}

\def\secref#1{Sec.~\ref{#1}}
\def\figref#1{Fig.~\ref{#1}}
\def\tabref#1{Tab.~\ref{#1}}
\def\eqref#1{Eq.~(\ref{#1})}

\usepackage{enumerate} 

\usepackage[dvipsnames,table]{xcolor} 
\usepackage{hhline}

\usepackage{hyperref}
\hypersetup{
	colorlinks=true,
	linkcolor=red,
	filecolor=black,      
	urlcolor=blue,
}
\usepackage{pifont}
\newcommand{\cmark}{\ding{51}}%
\newcommand{\xmark}{\ding{55}}%

\usepackage{gensymb}

\usepackage{todonotes}

\usepackage{subcaption}
\usepackage[font=footnotesize]{caption}

\usepackage{etoolbox}
\makeatletter
\patchcmd{\@makecaption}
{\scshape}
{}
{}
{}
\makeatletter
\patchcmd{\@makecaption}
{\\}
{:\ }
{}
{}
\makeatother

\usepackage{mathtools,xparse}
\DeclarePairedDelimiter{\abs}{\lvert}{\rvert}
\DeclarePairedDelimiter{\norm}{\lVert}{\rVert}

\let\labelindent\relax
\usepackage[inline]{enumitem}

\usepackage{multirow}

\title{\LARGE \bf Seeing the Wood for the Trees: Reliable Localization\\
	in Urban and Natural Environments}

\author{Georgi Tinchev, Simona Nobili and Maurice Fallon
	\thanks{The authors are with the Oxford Robotics Institute at the University of Oxford, United Kingdom. \texttt{\{gtinchev,snobili,mfallon\}@robots.ox.ac.uk}}%
	\thanks{S. Nobili is with the Institute of Perception, Action and Behaviour, School of Informatics, University of Edinburgh, UK.}%
	\thanks{This work was supported by the EPSRC RAIN and ORCA Robotics Hubs (EP/R026084/1 and EP/R026173/1 respectively). M. Fallon is supported by a Royal Society University Research Fellowship.}
}
\begin{document}
	\maketitle
	\thispagestyle{empty}
	\pagestyle{empty}
	
	\begin{abstract}
		In this work we introduce Natural Segmentation and Matching (NSM), an algorithm for reliable localization, using laser, in both urban and natural environments. Current state-of-the-art global approaches do not generalize well to structure-poor vegetated areas such as forests or orchards. In these environments clutter and perceptual aliasing prevents repeatable extraction of distinctive landmarks between different test runs. In natural forests, tree trunks are not distinctive, foliage intertwines and there is a complete lack of planar structure. In this paper we propose a method for place recognition which uses a more involved feature extraction process which is better suited to this type of environment. First, a feature extraction module segments stable and reliable object-sized segments from a point cloud despite the presence of heavy clutter or tree foliage. Second, repeatable oriented key poses are extracted and matched with a reliable shape descriptor using a Random Forest to estimate the current sensor's position within the target map.	We present qualitative and quantitative evaluation on three datasets from different environments - the KITTI benchmark, a parkland scene and a foliage-heavy forest. The experiments show how our approach can achieve place recognition in woodlands while also outperforming current state-of-the-art approaches in urban scenarios without specific tuning.
	\end{abstract}
	
	\section{Introduction}
	\label{sec:intro}
	
	
	Localization is an important problem in autonomous robot navigation when 
	surveying challenging environments such as forests, oil rigs, and disaster 
	response sites or teach-and-repeat operations in orchards for disease detection 
	and analysis of tree growth. Systems should be able to recognize places in both urban and vegetated areas, while being resilient to 
	appearance changes caused by the robot's motion, occluding clutter or temporal 
	variations of the environment.

	\begin{figure}[!ht]
		\centering
		\includegraphics[width=0.99\linewidth]{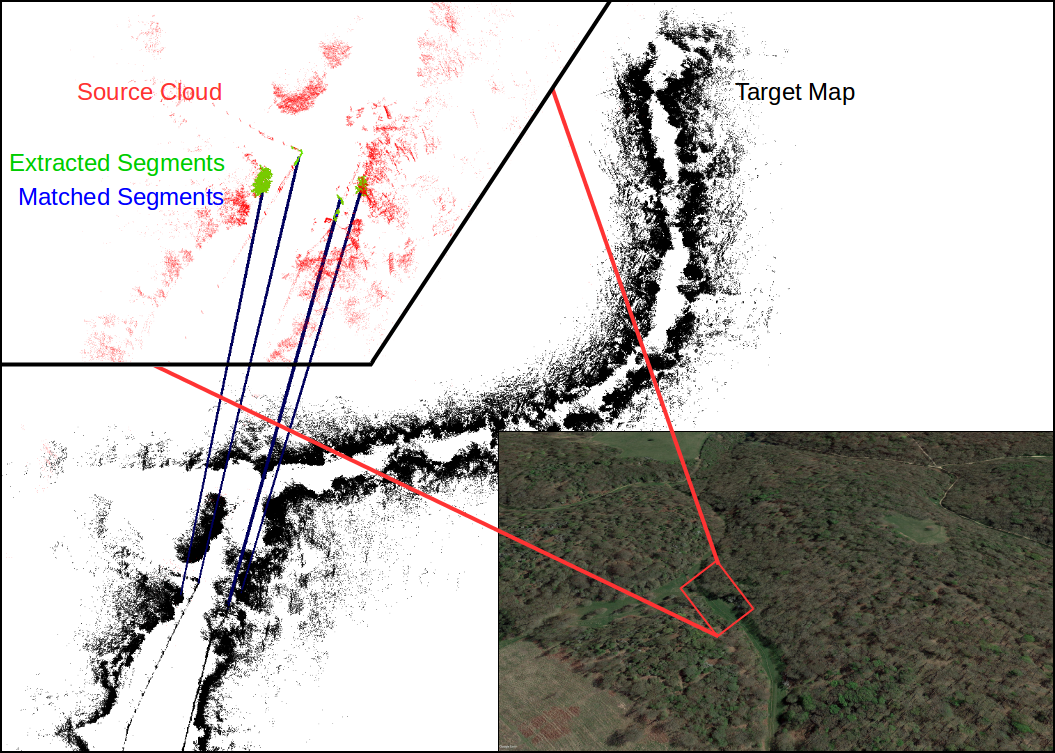}
		\caption{The proposed approach provides reliable segmentation (green) and 
			matching (blue lines) of objects such as trees and bushes within natural,  
			forested environments. The current \textit{source} point cloud (red)
			is shown in the top left corner. The blue lines 
			indicate the localization estimate relative to the \textit{target} map 
			(black), consisting of a previously traversed $\sim1\text{km}$ route within a 
			forest.}
		\label{fig:motivation}
		\vspace{-0.6cm}
	\end{figure}

	There has been extensive study of localization in indoor or outdoor structured 
	environments using vision or laser sensors. Strategies are often split into two 
	parts - global registration approaches which match high level features within 
	a large map~\cite{bosse2013place, dub2017icra, elbaz3d}, and fine registration methods that perform point-wise refinements~\cite{pomerleau13ar,SerafinNICP}. 
	These strategies often rely on landmark correspondences 
	between candidate point clouds, implicitly assume built environments, containing planar objects, or require reliable normal 
	estimation and consistent data association.
	
	The problem of place recognition in structure-poor scenarios is less well 
	explored. Current approaches model natural environments specifically, using  
	state machines to aid their place recognition systems based on prior knowledge 
	about the dataset~\cite{wellington2012orchard,underwood2015lidar}. It is common to specifically model the canopy of trees or 
	bushes. The foliage 
	in vegetated areas changes with vegetation growth or pruning, and is also 
	affected by continuous seasonal alteration. In addition, defining an accurate localization is difficult because the tree canopy can be interleaved and heavy foliage/clutter can obscure stable landmarks. 
	
	It is currently unclear how localization approaches will transfer to the more 
	challenging environments we are interested in. Designing algorithms to be invariant to the type of environment and geometrical differences may result in performance improvement. In 
	this work we bridge the gap between structured urban environments and 
	challenging natural environments without explicitly modelling every orchard or 
	forest individually by using more repeatable and descriptive features.
	
\begin{figure}[t]
		\vspace{+0.15cm}
		\centering
		\fbox{\includegraphics[width=0.95\linewidth]{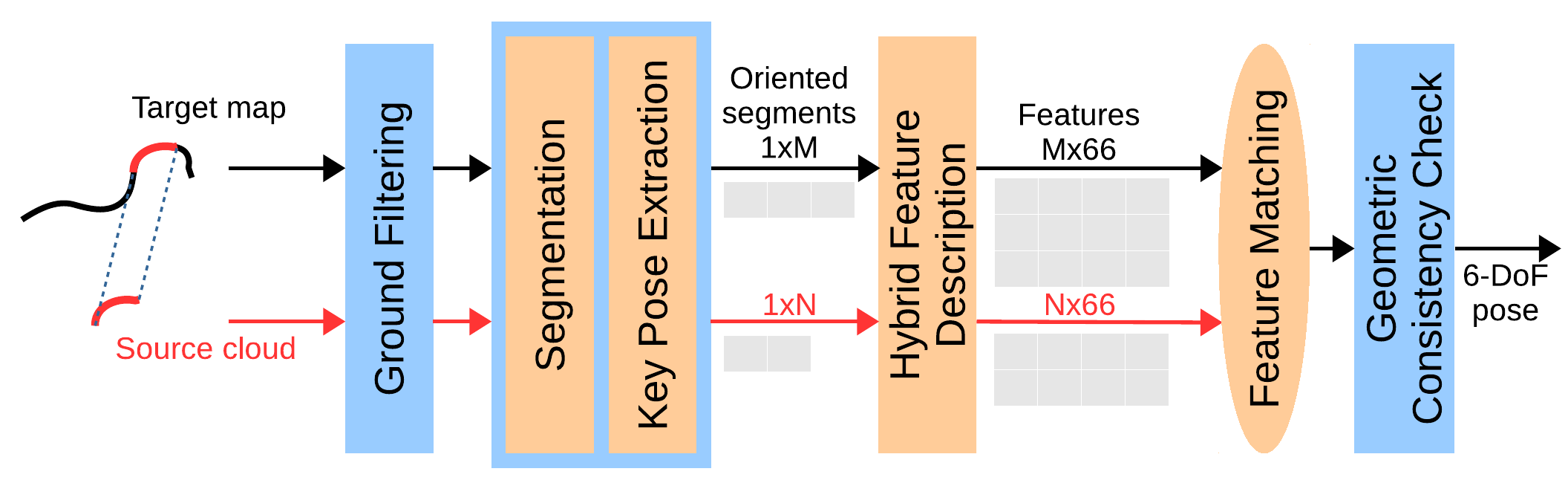}}
		\caption{The proposed approach.}
		\label{fig:nsm_approach}
		\vspace{-0.4cm}
	\end{figure}

	The main contribution of our work is a method that enables global place 
	recognition\footnote{In this system no initial estimate of the sensor's pose 
		is needed to initialize the alignment.} in three different scenery 
	settings. 
	Our approach adapts the framework from~\cite{dub2017icra} to
	reliably extract segments from natural environments which
	carry repeatable and distinctive information from a \textit{source} point cloud 
	and utilize them to solve global registration to a \textit{target} map 
	(\figref{fig:motivation}).  The pipeline of our approach is illustrated 
	in \figref{fig:nsm_approach}. 
	
	Our contributions are as follows:
	\begin{enumerate}
		\item we propose an extension of the SegMatch approach~\cite{dub2017icra} for place recognition in both urban and natural environments, which we call Natural Segmentation 
		and Matching (NSM),
		\item we identify and evaluate a novel combination of key pose extraction 
		and description methods. The key pose extraction module segments and defines consistent orientated coordinate frames for object-sized segments despite the presence of 
		heavy clutter. The descriptor carries sufficient 
		information to recognize different instances of the same segment, achieving 
		a higher accuracy with respect to previous approaches,
		\item we perform a thorough evaluation of the approach across datasets 
		captured in an urban area, a parkland scene and a foliage-heavy forest.
	\end{enumerate}

	The remaining sections of the paper are structured as follows: \secref{sec:related} presents the literature review,~\secref{sec:main} outlines each module of our algorithm, and~\secref{sec:exp} extensively evaluates the approach. We discuss interesting findings and limitations of our approach in~\secref{sec:discussion}.

	\section{Related Work}
	\label{sec:related}
	
	
	Global localization is commonly solved by directly extracting and describing 
	keypoints from a point 
	cloud~\cite{bosse2013place,bosse2009keypoint,bosse2010place}, 
	or by segmenting objects in the scene and matching those to a prior 
	map~\cite{dub2017icra,elbaz3d}. A coarse alignment is computed that can later 
	be refined with methods like the Iterative Closest Point (ICP)~\cite{Besl92pami}. The task is 
	more challenging in natural environments, as keypoint extraction or 
	segmentation methods often assume some regular structure in the environment.
	
	\subsection*{Place recognition strategies}
	
	Bosse and Zlot~\cite{bosse2009keypoint} presented three different methods for keypoint selection that were described with model grid descriptors. The descriptors' dimensions were later normalized using a nonlinear normalization function and reduced to increase efficiency and reduce the signal-to-noise ratio. Place recognition was achieved by keypoints \textit{voting} for their closest neighbour in a previous database of keypoints. Their work was extended in~\cite{bosse2010place} to support 3D regional point descriptors.
	
	Elbaz et al.~\cite{elbaz3d} utilized a Random Sphere Cover Set (RSCS) to divide 
	a point cloud into a set of super-points clusters. Each point can be a 
	part of multiple super-points, resulting in the point cloud being separated by 
	overlapping spheres. These spheres were projected as depth map images and fed into a 
	Deep Auto-Encoder network to generate feature descriptors. Candidate matches 
	are selected using a K-nearest neighbors (k-NN) search. In their approach when 
	segmenting the point clouds into RSCS, the authors assumed that each super-point described a surface.
	
	
	Dub{\'e} et al.~\cite{dub2017icra} presented SegMatch. This approach segmented
	the target and source point clouds 
	using Euclidean clustering. These clusters were matched using k-NN and 
	predictions from a Random Forest, trained on an Eigenvalue-based and Ensemble of 
	Shape Histogram features. The centroids of the matched segments were then used to produce a 6-DoF pose estimate for the sensing vehicle using RANSAC. The approach 
	assumes that the centroid (mean) of all points is a representative point of a 
	cluster, which unfortunately does not hold in natural environments. 
	
	\subsection*{Localization in Natural Environments}
	
	Some of the challenges in natural environments include occlusions, clutter or branch deformations, unstable normal extraction, predominance of structure-poor objects. These are usually the result of interleaved trees or even just windy conditions. Thus most state-of-the-art approaches performing 
	place recognition in natural environments 
	rely on explicit assumptions about the model of the scene~\cite{wellington2012orchard,underwood2015lidar}.
	
	Wellington et 
	al.~\cite{wellington2012orchard} presented an approach for tree segmentation in 
	orchards, utilizing a push-broom laser scanner. The ground surface was estimated 
	using a Markov Random Field, which helped estimating the boundaries between 
	each tree, the height and density of trees. A Hidden Semi-Markov Model modeled the environment using three states - tree, gap or boundary. The 
	approach incorporated a hidden state to account for the explicit prior on tree 
	spacing. Underwood et al.~\cite{underwood2015lidar} extended this approach to
	perform tree recognition and platform localization. After segmentation and 
	characterization, the localization problem was modelled as a Hidden Markov Model~\cite{rabiner1989tutorial}. The approach was tested in an orchard
	with regular rows of trees which were clearly delineated. 
	

	Bosse and Zlot~\cite{bosse2013place} performed place recognition in natural 
	environments by selecting a set of keypoints from 10\% of 
	the densest areas in a point cloud, and computed a local Gestalt descriptor 
	around each of them~\cite{bosse2009keypoint}.
	The method was evaluated on a natural open 
	eucalyptus forest with multi-use dirt trails, where the densest areas of the 
	point cloud were vegetated.
	In the proposed work, we utilize a different strategy for 
	keypoint extraction which is based on firstly extracting segments in the 
	environment, which we then append with a common orientation frame and describe 
	using a Gestalt descriptor. A notable difference is that we 
	opted for a learning-based approach when matching between key poses.
	
	In summary, our work differs from the state-of-the-art in place recognition in 
	natural environments in that we utilise segmentation as a prior step for 
	keyframe extraction on natural data. Our work adapts SegMatch~\cite{dub2017icra}
	to reliably 
	segment key poses in manner which is repeatable between 
	different observations. In addition, we utilise a feature descriptor that 
	describes both the shape of the segment and the corresponding key pose's 
	neighbourhood, resulting in resilient localization in cluttered natural areas 
	and urban scenarios.
	
	\begin{figure}[t]
		\vspace{+0.1cm}
		\centering
		\includegraphics[width=0.9\linewidth]{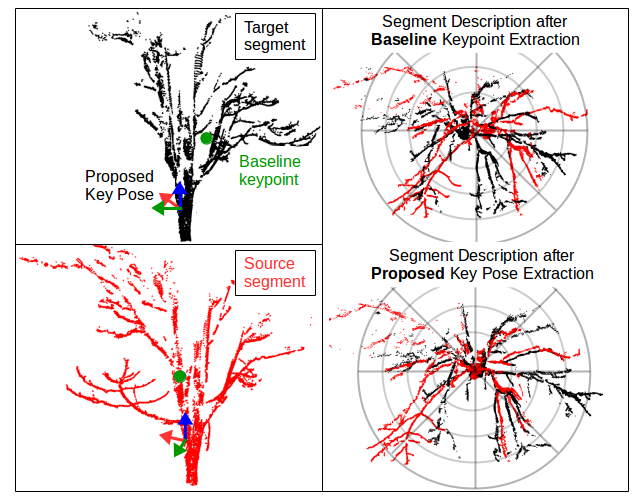}
		\caption{Key pose estimation between two different observations of the same 
			segment: the target map (black) and source cloud (red). The centroid of 
			each segment shown in green, used by a baseline 
			method~\cite{dub2017icra}. The proposed method computes position and 
			orientation for the key pose which is consistent for both the target and source clouds. The figures on the right represent a top view of the segments aligned with either of these two approaches. Subsequently our key pose descriptor can more coherently extract features as the points of each segment occupy the same descriptor bin in the two observations.}
		\label{fig:keypoint_extraction}
		\vspace{-0.7cm}
	\end{figure}
	\section{Natural Segmentation and Matching}
	\label{sec:main}
	
	This section presents the proposed Natural Segmentation and Matching (NSM) approach, which
	adapts SegMatch~\cite{dub2017icra} for natural environments. 
	We wish to estimate the position of the sensor within a target map using the current source point cloud. The approach focuses on 
	identifying repeatable segments from point clouds captured in cluttered scenes 
	where the detected objects can differ significantly in appearance between 
	observations. 
	We aim to overcome the weaknesses of the previous methods by improving the feature extraction and description steps as follows: \textit{A)} prior to 
	segmentation, the input clouds are pre-filtered using a 
	Progressive Morphological Filter to discard ground points 
	\cite{zhang2003progressive}. Then oriented key poses are computed for each segmented object. This makes the extraction more robust to variation in 
	appearance from different points of view. \textit{B)} A hybrid feature 
	descriptor is defined using PCA and Gestalt features, which embed sufficient 
	information so as to enable place recognition in both natural and urban environments.
	\vspace{-0.3cm}
	\subsection{Feature Extraction}
	\paragraph*{\textbf{Segmentation}}
	
	We employ Euclidean segmentation, similarly to~\cite{dub2017icra}, in order to delineate individual objects such 
	as interleaved trees and bushes. A typical segment corresponds to a tree with 
	a portion of its major branches, a dense bush, a vehicle, some 
	urban structure or any rigid object in the scene with limited physical size. 
	Most clutter is implicitly filtered-out as it does not satisfy criteria about the uniformity of points and dimensions of the segment in question. 
	Care is taken to ensure that it is possible to reliably distinguish a specific 
	object across temporal or spatial variations by assigning a key pose 
	to each segment. Note, that we considered the region-growing approach of~\cite{rabbani2006segmentation}, however, normals computed in these natural environments are unstable~\cite{bosse2013place}.\\

	\paragraph*{\textbf{Oriented Key Pose Extraction}} For each 
	segment in the source cloud ${s_s}_i \in \{{s_s}_1,{s_s}_2 \ldots {s_s}_N\}$, and in the target map ${s_t}_j \in \{{s_t}_1,{s_t}_2 \ldots {s_t}_M\}$, a 
	key pose $\mathbf{k} \in \mathbb{R}^6$ needs to be 
	defined. This pose will be used to define the position and orientation of each segment so as to aid the 
	recognition task. An illustration of how a consistent key pose can help mitigate 
	changes in appearance between different points of view is shown in 
	\figref{fig:keypoint_extraction}.
	
	We assume that the observed objects are rigidly connected to the ground. 
	This allows us to orient the \emph{z-axis} of $\mathbf{k}$ as the global up 
	vector, and determine the orientation of the segment based on the predominant dimension 
	along which the points are distributed perpendicularly to \emph{z}. The 
	\emph{x-axis} is projected onto the normal direction at $\mathbf{k}$, and the 
	\emph{y-axis} is computed as the cross product between the \emph{x} and 
	\emph{z} axes. 
	Once the orientation is been determined, the key pose's position $[{k}_x, {k}_y, 
	{k}_z]$ is calculated as the median of all the points of a subset of $s$ and 
	stored for later use. The median is used instead of the mean so as to mitigate segmentation differences between matching objects as shown on~\figref{fig:keypoint_extraction}
	
	\begin{figure}[t]
		\vspace{+0.1cm}
		\centering
		\includegraphics[width=0.5\linewidth]{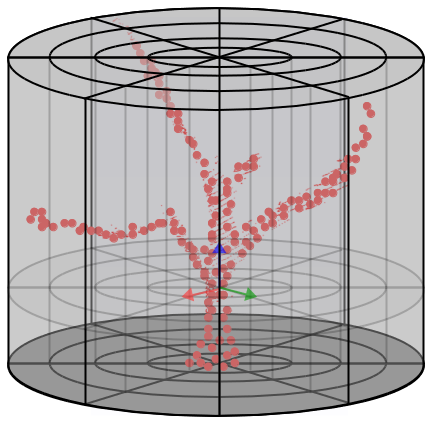}
		\caption{Visual representation of the Gestalt 
			descriptor of a tree segment (red), oriented 
			around the extracted key pose.}
		

		\label{fig:gestalt_descriptor}
		\vspace{-0.6cm}
	\end{figure}
	
	\subsection{Hybrid Feature Description}
	In the next step, the extracted segments are described in both the target and 
	source clouds using a hybrid feature set combining \textit{Eigenvalue} features and an adaptation of the 
	\textit{Gestalt} model~\cite{bosse2013place,walthelm2004enhancing}, illustrated 
	in \figref{fig:gestalt_descriptor}. Thus, NSM encodes both the geometric properties of the segments and their point distribution.
	
	For each segment $s$, planarity and cylindricality features are computed 
	after PCA. These are, respectively,
	\begin{equation}
	P_{\lambda} = \lambda_2 - \lambda_1
	\end{equation}
	and
	\begin{equation}
	C_{\lambda} = \lambda_3 - \lambda_2,
	\end{equation}
	where $\lambda_{\{1,2,3\}}$ are to the ordered normalized 
	eigenvalues. The structure tensor is of rank 3 and it follows that 
	$\lambda_1 \geq \lambda_2 \geq \lambda_3 \geq 0$. $P_{\lambda}$ and 
	$C_{\lambda}$ encode the geometric information for each cluster, 
	enabling NSM to account for the rigidity of objects. 
	
	Each segment is then split into 4 radial and 8 azimuthal divisions about its 
	key pose's \emph{z-axis} and clock-wise starting from the \emph{x-axis}, resulting in 32 
	bins. For each bin the mean and variance of the height values of all points are 
	recorded for a total of 64 dimensions. The planarity and cylindricality values 
	are included in the feature vector, resulting in a descriptor of 66 dimensions. 
	In the experimental section we show that such a descriptor carries sufficient 
	information to identify different instances of the same segment in cluttered 
	scenarios.
	
	
	\subsection{Feature Matching}
	Feature matching is performed in two stages using the approach proposed 
	by~\cite{dub2017icra}. Firstly, a k-NN search in feature 
	space creates a list of possible matches for each source segment $s_{s_i}$ that 
	matches $K$ target segments using $L_2$ distance. Secondly, the list of 
	candidate matches is 
	further refined using a Random Forest (RF) that performs a binary classification 
	over each set of $K$ proposals. 
	
	We have trained our RF using the 
	full feature vector dimension while taking into consideration the difference 
	between two matching samples. The input to our RF consists of 
	$(|\mathbf{f}_{s}|, |\mathbf{f}_{t}|, 
	\Delta \mathbf{f})$, where $\mathbf{f}_s$ and 
	$\mathbf{f}_t$ are the feature vectors corresponding to all source segments and 
	its candidate matches in the target map and $|\cdot|$ denotes the modulus. We consider the descriptor $\Delta 
	\textbf{f}$ to be computed from 
	the difference between each source and target feature vector, that is $\Delta 
	\textbf{f}_{ik}=|\textbf{f}_{s_i} - \textbf{f}_{t_k}|$ $\forall{i,k}$ with $k 
	\in [1:K]$, which we empirically
	found to 
	help during the matching task. This is possible as the segments have a consistent orientation after key pose extraction and each feature vector is 
	sorted by definition. The output of the RF classifier is a score $w$, defining the likelihood of a match being correct between segments $s_s$ and $s_t$. The score is 
	thresholded to produce a set of accepted matches.
	
		\begin{table}[b]
		\centering
		\resizebox{\columnwidth}{!}{
			\begin{tabular}{|l|l|l|l|} 
				\hline
				& \multicolumn{1}{c|}{\textbf{KITTI}} & 
				\multicolumn{1}{c|}{\textbf{George Square Park (GS)}} & 
				\multicolumn{1}{c|}{\textbf{Cornbury Park (CP)}} \\ \hline
				\cellcolor{gray!25}Source & Geiger et al.~\cite{geiger2013vision} & 
				Ours & Ours \\
				\hline
				\cellcolor{gray!25}Environment & Urban, structured & Vegetated, 
				structured, & Vegetated, unstructured,\\
				\cellcolor{gray!25} &  & well observed trees & hard to delineate 
				trees, \\
				\cellcolor{gray!25} &  &  & predominance of bushes\\
				\hline
				\cellcolor{gray!25}Dynamics & Moving cars, bicycles, & A few moving 
				people & None\\
				\cellcolor{gray!25} & pedestrians &  & \\
				\hline
				\cellcolor{gray!25}LIDAR & 3D HDL-64E @ 10Hz & Push-broom @ 75Hz & 
				3D HDL-32E @ 10Hz \\
				\hline
				\cellcolor{gray!25}Path & On streets & 3 loops on a concrete  & 
				Off-road dirt track\\
				\cellcolor{gray!25} &  & trail in the park & out and back trail
				\\
				\hline
				\cellcolor{gray!25}Scene Area & $\sim 500\text{m}$ for localization & 
				$\sim 500\text{m}$ each loop & $\sim 1\text{km}$ each direction\\
				\hline
				\cellcolor{gray!25}\# of Scans & 1104 & $\sim$ 40700 & 1823\\
				\hline
				\cellcolor{gray!25}\# of Points & $\sim$ 130000/scan & $\sim$ 
				381/scan & $\sim$ 70000/scan\\
				\hline
				\cellcolor{gray!25}Map creation & Scans projected & 
				VO~\cite{huang2017visual} on first loop,& Scans projected on 
				ground\\
				\cellcolor{gray!25} & on ground truth &manual loop 
				closures~\cite{kaess2008isam} & truth from forward pass \\			
				\cellcolor{gray!25} &  & & locally corrected via ICP\\
				\cellcolor{gray!25} &  & & globally corrected 
				via~\cite{kaess2008isam}\\
				\hline
				\cellcolor{gray!25}Ground Truth & \multicolumn{1}{c|}{\cmark} & 
				\multicolumn{1}{c|}{\xmark} & \multicolumn{1}{c|}{\cmark}\\
				\hline
				\cellcolor{gray!25}Training Data & \multicolumn{1}{c|}{KITTI Seq 
					06} & 
				\multicolumn{1}{c|}{Pre-trained KITTI} & 
				\multicolumn{1}{c|}{Pre-trained KITTI}\\
				\cellcolor{gray!25}Testing Data & \multicolumn{1}{c|}{ KITTI Seq 
					00} & 
				\multicolumn{1}{c|}{Georges Square (GS)} & 
				\multicolumn{1}{c|}{Cornbury Park (CP)}\\
				\hline
				\cellcolor{gray!25}Experiment & \multicolumn{1}{c|}{Exp. A} & 
				\multicolumn{1}{c|}{Exp. B} & \multicolumn{1}{c|}{Exp. C} \\
				\hline
		\end{tabular}}
		\caption{Datasets used in our experiments.}
		\label{table:datasets}
	\end{table}

	\subsection{Pose Estimation}
	We utilize the method from~\cite{aldoma2012global,dub2017icra} in order to maintain 
	geometric consistency between each pair of matches 
	$((\textbf{k}_{s_p},\textbf{k}_{t_p}),(\textbf{k}_{s_q},\textbf{k}_{t_q}))
	\in \mathcal{C} = \{(\textbf{k}_{s_0},\textbf{k}_{t_0}) \ldots 
	(\textbf{k}_{s_L},\textbf{k}_{t_L})\}$, where $\mathcal{C}$ is 
	the set of accepted matches proposed by the RF and $L$ its cardinality. The 
	formula in~\eqref{eq:gc} ensures that a pair of matching key poses have 
	similar spatial distances between them.

		\begin{equation}
		\abs{\norm{\textbf{k}_{t_p} - \textbf{k}_{t_q}}_2 - \norm{\textbf{k}_{s_p} - 
				\textbf{k}_{s_q}}_2} < \epsilon
		\label{eq:gc}
		\end{equation}


		In~\eqref{eq:gc}, $\epsilon$ is the resolution parameter, which determines how 
		similar the source cloud is to the target. The lower the value, 
		the stricter the algorithm behaves. A localization proposal is accepted only if 
		it satisfies a criteria about the minimum number of 
		accepted matches. For a 6-DoF pose to 
		be estimated, there must be at least $\tau=3$ matches.
		Finally the 6-DoF transform which aligns the source cloud into the target map is 
		computed by a RANSAC optimization on the key poses of the remaining matches.

		\begin{table}[t]
			\vspace{+0.15cm}
			\centering
			\resizebox{1.0\columnwidth}{!}{
				\begin{tabular}{|l|l|l|l|l|} 
					\hline
					\multicolumn{5}{|c|}{\textbf{Parameters}} \\
					\hline
					\multicolumn{2}{|c|}{} & \multicolumn{1}{c|}{\textbf{KITTI}} & 
					\multicolumn{1}{c|}{\textbf{GS}} & 
					\multicolumn{1}{c|}{\textbf{CP}} \\
					\hline
					\cellcolor{gray!25} & \cellcolor{gray!25}Min 
					\# Points & 
					\multicolumn{2}{c|}{$200$} & 
					\multicolumn{1}{c|}{$2500$}\\
					\cellcolor{gray!25} & \cellcolor{gray!25}Max 
					\# Points & 
					\multicolumn{2}{c|}{$1500$} & 
					\multicolumn{1}{c|}{$50000$}\\
					\cellcolor{gray!25}{\multirow{-3}{*}{\textbf{Segmentation}}} & 
					\cellcolor{gray!25}Max Distance & 
					\multicolumn{3}{c|}{$0.2\text{m}$} \\
					\hhline{--}
					\cellcolor{gray!25}{\textbf{Description}} & 
					\cellcolor{gray!25}Gestalt 
					Radius & 
					\multicolumn{3}{c|}{$2.0\text{m}$} 
					\\
					\hhline{--}
					\cellcolor{gray!25} & \cellcolor{gray!25}K 
					Neighbours & \multicolumn{3}{c|}{$200$} \\
					\cellcolor{gray!25} & \cellcolor{gray!25}\# of Trees & 
					\multicolumn{3}{c|}{$250$} \\
					\cellcolor{gray!25} & \cellcolor{gray!25}Tree Depth & 
					\multicolumn{3}{c|}{$50$} \\
					\cellcolor{gray!25}{\multirow{-4}{*}{\textbf{Matching}}} & 
					\cellcolor{gray!25}RF Threshold ($w$) & 
					\multicolumn{3}{c|}{$0.69$} \\
					\hhline{--}
					\cellcolor{gray!25}{\textbf{Pose}} & \cellcolor{gray!25}Resolution 
					($\epsilon$) & 
					\multicolumn{3}{c|}{$0.4\text{m}$} \\
					\cellcolor{gray!25}{\textbf{Estimation}} & \cellcolor{gray!25}Min 
					\# Clusters ($\tau$) & 
					\multicolumn{3}{c|}{$4$} \\ \hline
				\end{tabular}
			}
			\caption{Parameters used in our experiments.}
			\label{table:params}
			\vspace{-0.3cm}
		\end{table}
		
		\section{Experimental Evaluation}
		\label{sec:exp}
		
		\begin{figure}[b]
			\centering
			\includegraphics[height=1.5in]{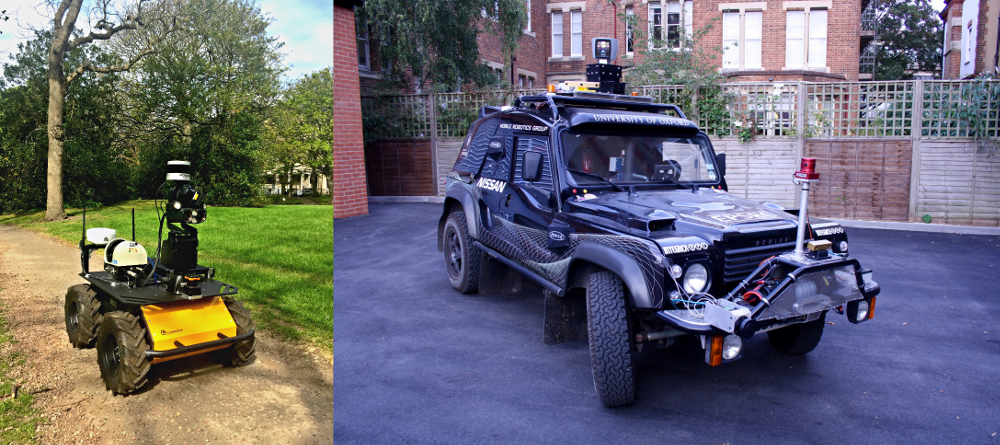}
			\caption{The robots utilised to collect data for our experiments. A Clearpath Husky robot (left) was used to collect the GS dataset with a push-broom LIDAR and a Multisense SL. A Bowler Wildcat platform (right) was utilised to collect the CP dataset, equipped with a Velodyne HDL-32E LIDAR, OxTS GPS, Novatel INS and Bumblebee2 camera.}
			\label{fig:robots}
		\end{figure}
		
		The main goal of this work is to perform global place recognition relative to a prior LIDAR map, without any prior information about the sensor's current pose. Our experiments are 
		designed to demonstrate the capabilities of our system and are performed on three 
		different datasets with increasing complexity, as shown on~\tabref{table:datasets}. Their corresponding parameters are presented in~\tabref{table:params}. 
		
		The RF used in all of our experiments was trained on Seq. 06 of the KITTI dataset, using a careful manual annotation of matching pairs of segments between individual clouds. We identified 487 true 
		matches (distance between segments $<0.5\,m$) and 218513 negative segment correspondences (distance $\ge 0.5\,m$) in this manner.
		
		The proposed approach was tested against SegMatch~\cite{dub2017icra} in all experiments, which is indicated as a baseline where appropriate. Both NSM and SegMatch produce localization estimates relative to a prior map, for every 1 meter of vehicle distance travelled. 
		
		To evaluate the accuracy, the estimated pose $T_e$, was compared to the best ground truth pose we could achieve, $T_c$. The error $\Delta T$ is computed as follows:

		\begin{equation}
		\Delta T = 
		\begin{bmatrix}
		\Delta \mathbf{R}       &  \Delta \mathbf{t} \\
		0     & 1 
		\end{bmatrix}
		= T_eT_c^{-1}
		\label{eq:error_metric}
		\end{equation}
		
		where $\Delta \mathbf{t}$ represents the translation error and $\Delta \mathbf{R}$ the rotation error in matrix form. The 3D translation error $e_t$ is defined as the Euclidean distance of the translation vector $\Delta \textbf{t}$ as follows:
		
		\begin{equation}
		e_t = ||\Delta \mathbf{t}|| = \sqrt{\Delta x^2 + \Delta y^2 + \Delta z^2}
		\end{equation}
		
		The 3D rotation error $e_r$ is defined as the Geodesic distance given the rotation error $\Delta \mathbf{R}$ as follows:
		\begin{equation}
		e_r = \arccos \Big( \frac{\text{trace}(\Delta \mathbf{R})-1}{2} \Big)
		\end{equation}

		\begin{figure}[t]
			\vspace{-0.5cm}
			\centering
			\includegraphics[width=\linewidth]{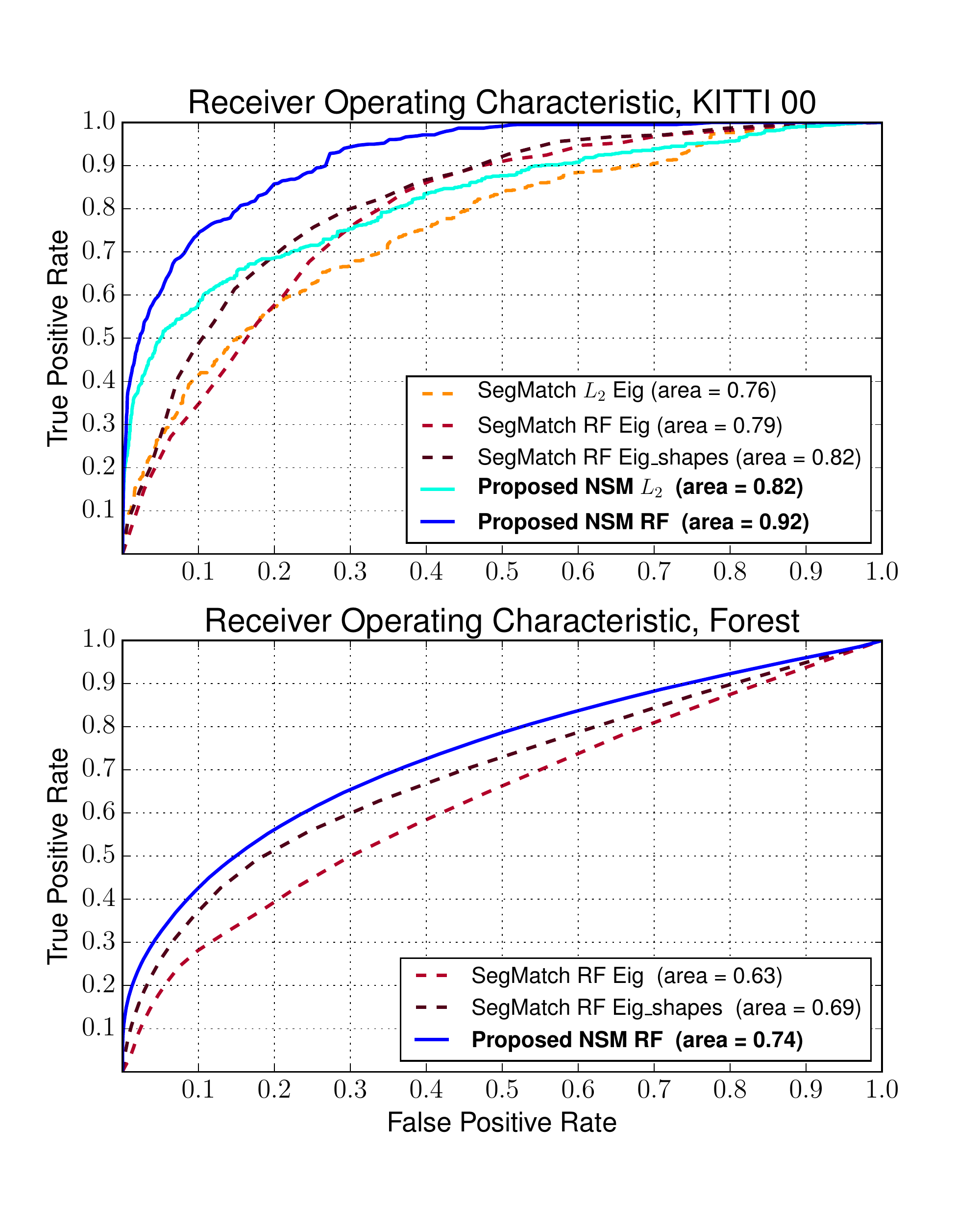}
			\vspace{-1.1cm}
			\caption{Receiver Operator Characteristic curves for NSM and SegMatch~\cite{dub2017icra} evaluated on Seq 00 of the KITTI dataset (top) and a forest dataset, similar to GS (bottom). The proposed approach was evaluated with both k-NN in feature space ($L_2$, cyan) and a pre-trained Random Forest (RF) classifiers (blue). The baseline was evaluated with Eigenvalue-based (Eig) features and Eigenvalue with ESH (Eig\_shapes) features.}
			\label{fig:roc_segMatch_vs_nsm}
			\vspace{-0.7cm}
		\end{figure}

		The choice of evaluation metric is motivated by~\cite{pomerleau13ar}.
		
		The following three experiments were carried out to support our claims:
		
		\begin{enumerate}[label=\Alph*)]
			\item Quantitative analysis of the performance of the proposed hybrid feature descriptor on urban and forested data with comparison to the baseline.
			\item Both qualitative and quantitative evaluation of our method performing in parkland environment, showing that the extracted segments are better represented by the oriented key pose resulting in more accurate localizations.
			\item A demonstration of the performance of our localization system in a very heavily vegetated environment.
		\end{enumerate}
		
		A video to accompany this paper is available at\\ \url{http://ori.ox.ac.uk/nsm-localization}.

		\begin{figure}[t]
			\vspace{+0.3cm}
			\centering
			\includegraphics[width=0.99                     \linewidth]{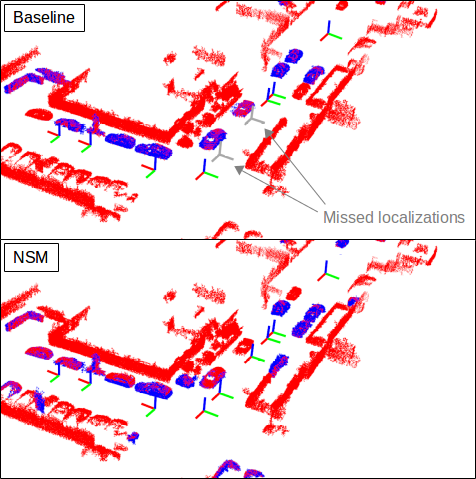}
			\caption{Qualitative illustration of the performance of our algorithm (bottom) on the KITTI dataset in comparison to the baseline (top). Blue point clouds represent correctly identified segments in the source cloud (red). Axis-colored poses indicate successful localizations, while grey colored are missed localizations.}
			\label{fig:matches_kitti}
			\vspace{-0.4cm}
		\end{figure}
		
		\subsection{Feature Description Analysis}

		The first experiment examines the performance of our hybrid feature descriptor. We use an approach similar to that presented in~\cite{dub2017icra}. We use the KITTI dataset and a forested environment similar to the GS dataset. The KITTI dataset was captured by a 3D Velodyne HDL-64E LIDAR mounted on a car in an urban environment. The forested dataset was recorded with a 3D Velodyne VLP-16 mounted on top of a Clearpath Robotics Husky UGV, as shown in~\figref{fig:robots}.
		
		~\figref{fig:roc_segMatch_vs_nsm} (top) 
		illustrates the Receiver Operating Characteristic (ROC) curves for three combinations of features and two classifiers. Namely, we compared Eigenvalue-based features and Ensemble of Shape Histogram (ESH) features, as proposed in~\cite{dub2017icra} to our hybrid feature extractor. Furthermore, we compared the performance of a RF classifier trained on the aforementioned features to the $L_2$ distance in feature space (k-NN). Our proposed approach extracted more positive matches
		than the baseline with both $L_2$ distance in feature space (cyan in comparison to dashed orange) and a RF classifier (blue compared to dashed brown), while limiting false positive correspondences. In the illustrative example on~\figref{fig:matches_kitti}, we show that the proposed approach was able to produce two more localizations as it successfully matched partial observations of segments from the front on the vehicle (bottom of~\figref{fig:matches_kitti}). 
		
		
		~\figref{fig:roc_segMatch_vs_nsm} (bottom) illustrates the ROC curve of the proposed features' classifier against the baseline on the forest dataset. The performance of the all the classifiers is significantly lower than on the urban dataset. The proposed feature extraction method has larger area under the curve (AUC) compared to the baseline. This result could be improved by learning features, optimized for specific environments. We plan to explore this topic in our future work.
		
		Based on these results, we chose a false positive rate of 0.1 (RF score $w=0.69$) for our subsequent experiments in order to 
		limit false positive segment matches and maximize performance.

		\begin{figure}[t]
			\centering
			\includegraphics[width=\linewidth]{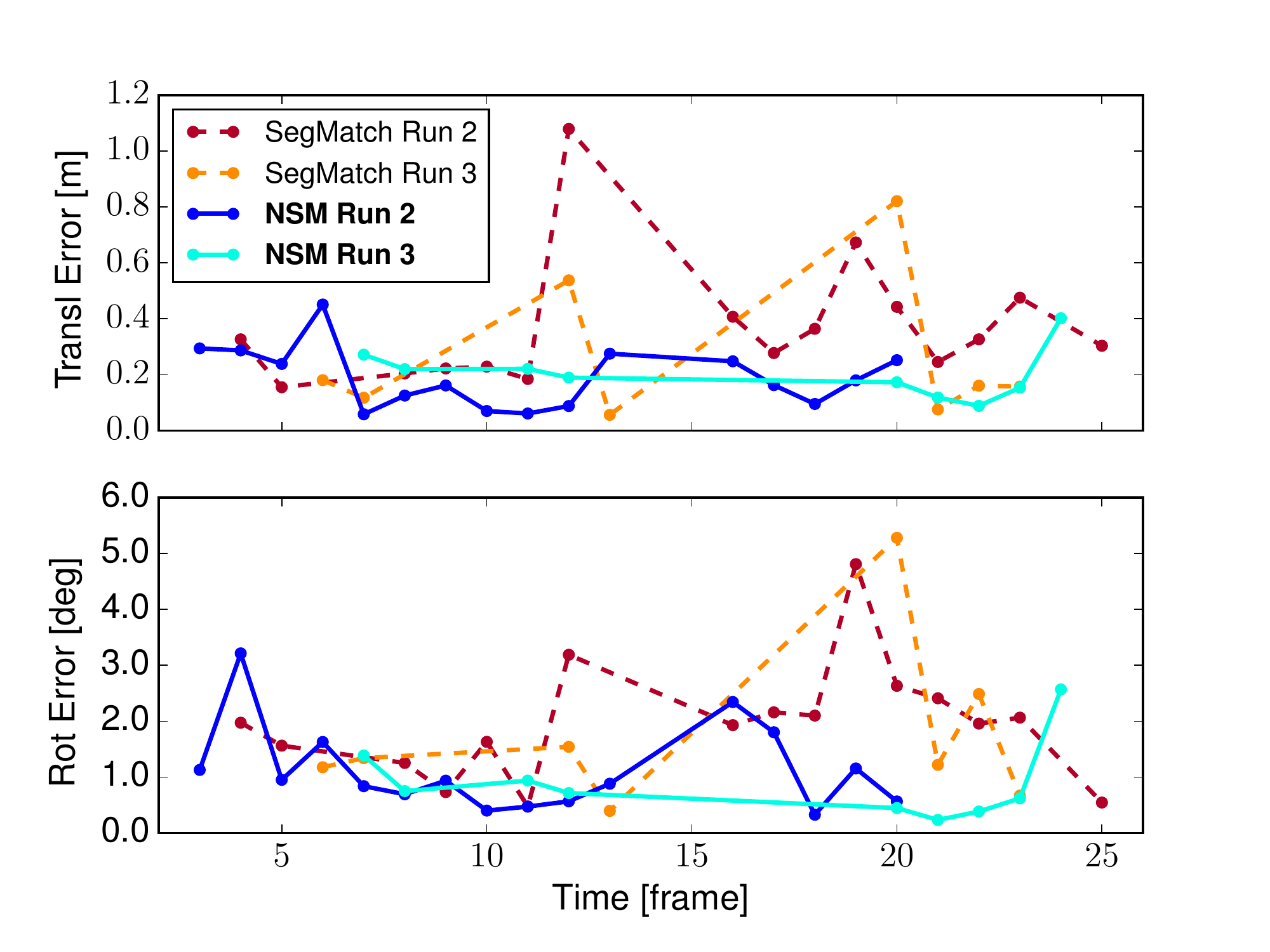}
			\caption{Translation (top) and rotation (bottom) errors for the proposed approach (continuous lines) and the baseline algorithm (dashed lines) tested on the GS dataset from two sequential runs. Run 1 was used to create the map.}
			\label{fig:nsm_vs_sm_gs}
		\end{figure}

		\subsection{Performance in a Parkland Environment}
		
		\begin{figure*}[t]
			\vspace{+0.2cm}
			\centering
			\begin{subfigure}[b]{0.5\textwidth}
				\centering
				\includegraphics[trim={0.5cm 0 0.5cm		
					0},clip,height=0.48\textwidth]{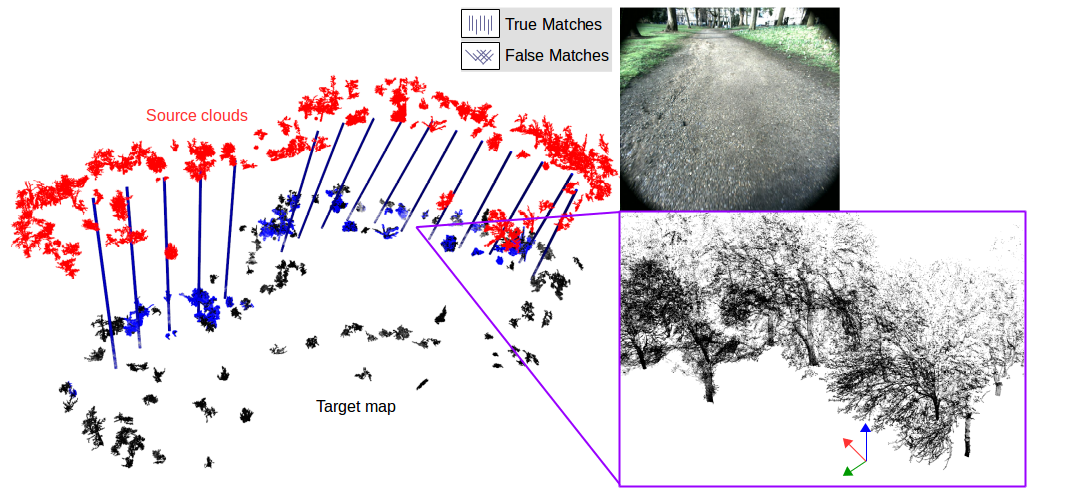}
			\end{subfigure}\hspace{-0.1cm}%
			\begin{subfigure}[b]{0.49\textwidth}
				\centering
				\includegraphics[trim={0 0 0 
					0},clip,height=0.46\textwidth]{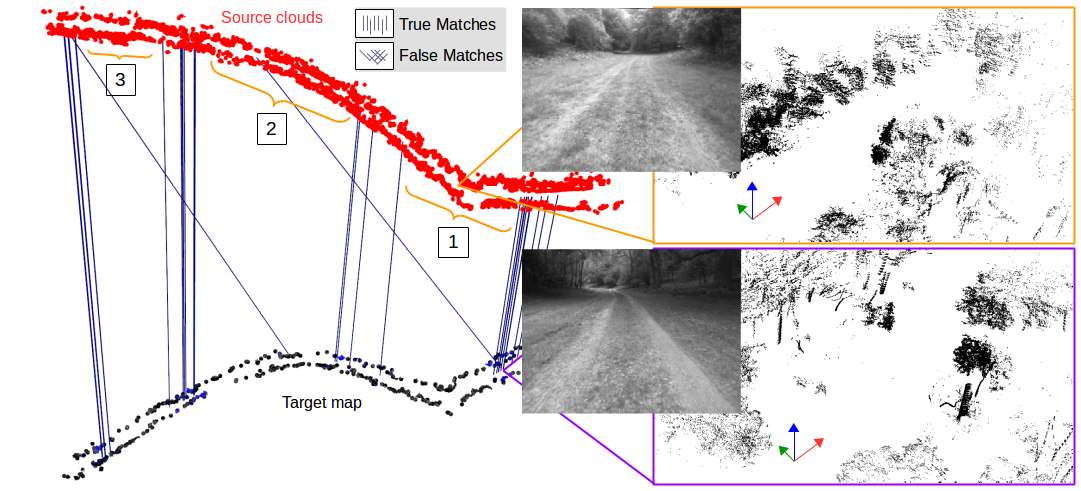}
			\end{subfigure}
			\caption{Illustration of the proposed approach on the GS dataset (left) and CP dataset (right). In both environments the source clouds (red) are being registered to an \textit{a-priori} target map (black). An example of a point cloud during successful localization in the presence of trees is shown in purple, while an example of a missed localization (observed in 1, 2, 3) in the absence of rigid objects is illustrated in orange. Images from the robot's point of view are shown to illustrate the challenging scenario.}
			\label{fig:results_cornbury_gs}
			\vspace{-0.4cm}
		\end{figure*}
		
		\begin{table}[t]
			\centering
			\resizebox{\columnwidth}{!}{
				\begin{tabular}{|c|l|l|l|l|l|}
					\hline
					\multicolumn{2}{|c|}{} & 
					\multicolumn{1}{c|}{\textbf{Transl RMSE}} & 
					\multicolumn{1}{c|}{\textbf{Rot 
							RMSE}} & 
					\multicolumn{1}{c|}{\textbf{Localizations}} & 
					\multicolumn{1}{c|}{\textbf{Frames}} \\ 
					\hline
					\cellcolor{gray!25} & Run 2                    & 
					$0.43\text{m} \pm 0.22\text{m}$     & $2.22^{\circ} \pm 1.03^{\circ}$      & 
					\multicolumn{1}{c|}{$\mathbf{16}$}                                  
					
					& 
					\multicolumn{1}{c|}{${26}$}                                
					\\ 
					\cline{2-2} 
					\cellcolor{gray!25}{\multirow{-2}{*}{\textbf{SegMatch}}} & Run 
					3                    & $0.37\text{m} \pm 0.25\text{m}$       & $2.28^{\circ} 
					\pm 1.45^{\circ}$    & 
					\multicolumn{1}{c|}{$8$}                                   
					
					& 
					\multicolumn{1}{c|}{${29}$}                                
					\\ 
					\hline
					\cellcolor{gray!25}      & Run 
					2                    & 
					$\mathbf{0.22m} \pm \mathbf{0.11m}$     & 
					$\mathbf{1.35}^{\circ} \pm \mathbf{0.76}^{\circ}$      & 
					\multicolumn{1}{c|}{$\mathbf{16}$}                                  
					
					& 
					\multicolumn{1}{c|}{${26}$}                                
					\\ 
					\cline{2-2} 
					\cellcolor{gray!25}{\multirow{-2}{*}{\textbf{NSM}}} & Run 
					3                    & 
					$\mathbf{0.22m} 
					\pm \mathbf{0.09m}$       & $\mathbf{1.12}^{\circ} \pm 
					\mathbf{0.67}^{\circ}$   & 
					\multicolumn{1}{c|}{$\mathbf{9}$}                                   
					
					& 
					\multicolumn{1}{c|}{${29}$}                                
					\\ \hline
				\end{tabular}
			}
			\caption{Quantitative results of the proposed approach and the baseline on GS dataset.}
			\label{tab:gs_results}
			\vspace{-0.5cm}
		\end{table}

		In a second experiment we aim to showcase the proposed key pose selection strategy of our system in a park environment. It consists of well-observed trees, benches and bushes spatially separated and with few interleaved branches between the trees.
		
		This experiment was carried out in George Square Park, Edinburgh, with a Clearpath Husky robot (\figref{fig:robots}, left) traversing three loops of $\sim500\text{m}$ each. The first loop was used to create the map (Run 1), while loops two and three (Run 2, 3) were used to test the localization. The vehicle was equipped with a push-broom LIDAR. Visual odometry~\cite{huang2017visual} was used to create 3D LIDAR swathes for both the prior map and the source clouds. The prior map was further corrected in a SLAM system using~\cite{kaess2008isam} by carefully identifying loop closures. For the source cloud, LIDAR swathes were accumulated every $\sim10\text{m}$ that the vehicle travelled. To account for the different sensor classes the segmentation algorithm parameters were changed as listed in~\tabref{table:params}. In this experiment we compared NSM to the baseline using centroid computation, Eigenvalue features and the RF classifier. Note that we tried utilising Eigenvalue+ESH features with a pre-trained RF, however, no localization proposals were produced with this system configuration.
		
		~\figref{fig:nsm_vs_sm_gs} illustrates translation and rotation error as computed using~\eqref{eq:error_metric} from the last two test runs. A visual representation of the environment and a running example of it are shown on~\figref{fig:results_cornbury_gs} (left). Neither the baseline, nor the proposed approach produced any false localizations. Quantitative results of the experiment are shown in~\tabref{tab:gs_results}. The proposed approach shows better accuracy for localization. We speculate that in the case of NSM the oriented key pose extraction strategy contributes for a closer match to the ground truth estimate, as previously illustrated on~\figref{fig:keypoint_extraction}.

		\subsection{Performance in a Foliage-Heavy Forest}

		The last experiment demonstrates the potential of our system to localize in highly vegetated scenes. The data for the experiment was collected in Cornbury Park, Oxfordshire, while traversing an approximately $1\text{km}$ path through woodland. The data was captured by a Bowler Wildcat equipped with a Velodyne HDL-32E LIDAR, producing 3D point cloud scans at 10Hz (\figref{fig:robots}, right). Pictures from a camera mounted on a robot in this environment are shown in~\figref{fig:results_cornbury_gs} (right). This environment was very challenging to localize in, even for a person manually comparing the prior map to a source point cloud. It consists of heavily interleaved foliage, trees and bushes without clear spatial separation. 
		
		\begin{figure}[t]
			\centering
			\includegraphics[width=\linewidth]{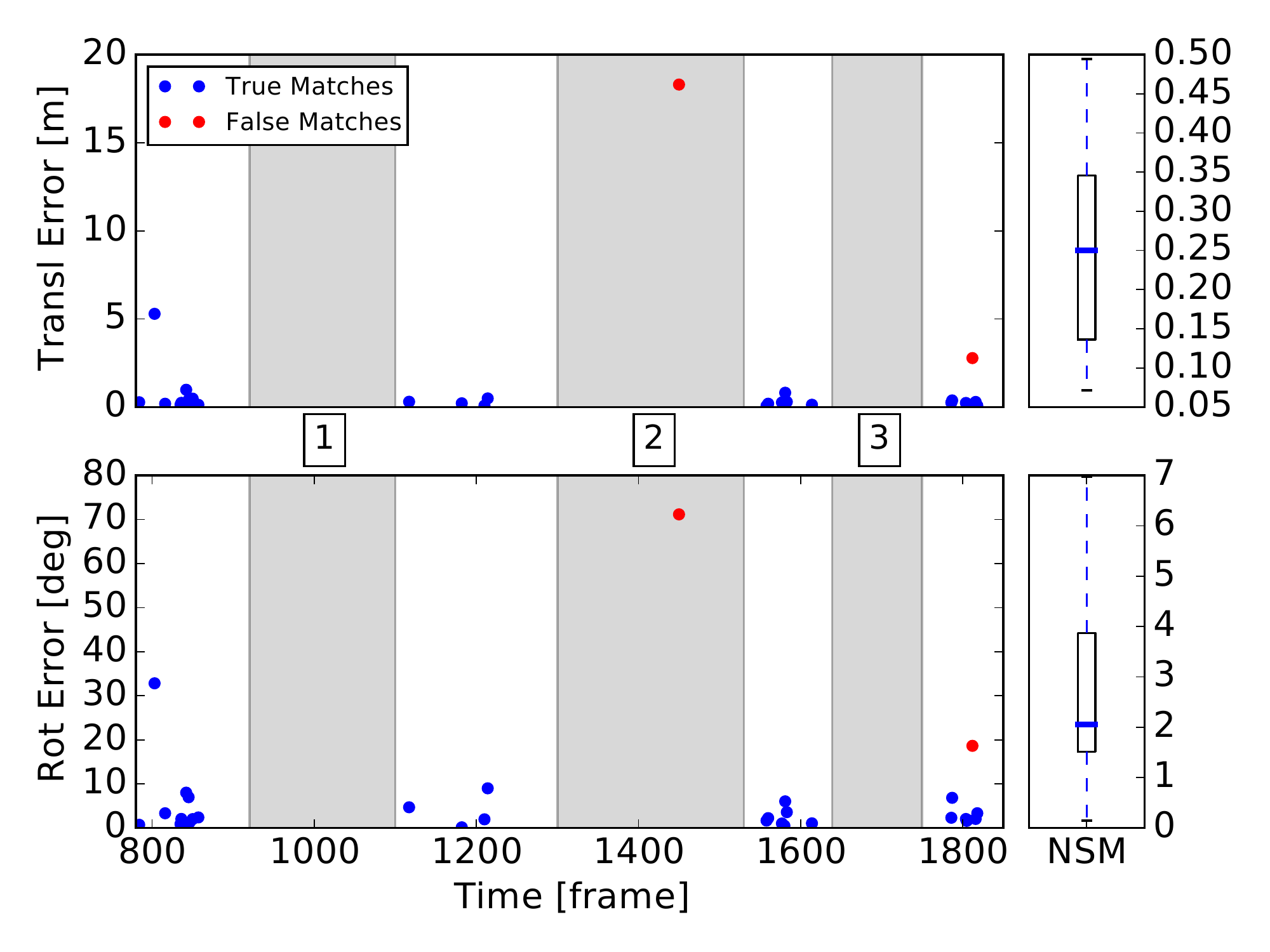}
			\caption{3D translation error (top) and rotation error (bottom) of correct and incorrect localizations, evaluated on the CP dataset. True matches (blue) correspond to the consistent vertical lines in~\figref{fig:results_cornbury_gs} (right), while incorrect localizations (red), correspond to inconsistent diagonal lines in~\figref{fig:results_cornbury_gs} (right). Boxplots on the right indicate the quantiles for the distribution of 0.50 (thick blue bars), 0.25 and 0.75 (lower and upper black rectangles) of true localization matches. During the experiment the vehicle traversed areas of continuous vegetation (1, 2, 3).}
			\label{fig:nsm_vs_icp_map_cornbury}
			\vspace{-0.5cm}
		\end{figure}
		
		Because the vehicle was equipped with a combined GPS/INS sensor, we intended to use this for evaluation. However, the route along the path combined periods of GPS reception and periods where trees covered the path and blocked GPS reception. Instead, we created the map by combining an initial vehicle trajectory, made by registering scans spaced every $30\text{m}$ to one another using ICP, with carefully chosen and reliable GPS measurements. To do this we smoothed the initial trajectory with the GPS measurements using~\cite{kaess2008isam} to achieve a best estimate of ground truth. Source point clouds were created when traversing the environment in the opposite direction by accumulating point clouds for every $\sim1\,\text{m}$ travelled. 
		
		~\figref{fig:nsm_vs_icp_map_cornbury} presents the translation and rotation errors of correct and incorrect localization matches, computed using~\eqref{eq:error_metric}. In the environment we discovered three sections of uninterrupted bushes obscuring any trees (marked with 1, 2, 3 on~\figref{fig:nsm_vs_icp_map_cornbury} and~\figref{fig:results_cornbury_gs}, right). In these areas we did not expect the algorithm to function, as the continuous hedge was visually challenging even for a person to distinguish. Outside these areas there were $96.7\%$ positive and $3.32\%$ false localization matches. We envisage that the two false detections were due to the radius of the Gestalt feature, as this feature's descriptiveness depends on the size and symmetry of the objects. In those specific cases two small bushes, with boundary points of their convex hull equal to the Gestalt radius, were incorrectly matched to bigger bushes that fully encapsulated them.
		
		In this environment the descriptiveness of the features enabled the proposed approach to produce a series of successful localizations. This was achieved by identifying partially observed and asymmetrical segments.
		
		\section{Discussion and Limitations}
		\label{sec:discussion}
		
		As described above the proposed approach is able to provide reliable localization proposals in urban and natural environments without specific parameter tuning, despite encountering challenging vegetated scenes. The current limitations of the algorithm are:
		
		\begin{enumerate}
			\item The segmentation of large objects can differ between the target map and the current point cloud when the sensor vantage point changes significantly. This would result in dissimilar segments which could not be matched. 
			\item The Gestalt descriptor has bins of varying sizes and the descriptor's azimuthal divisions intersect near the key pose. Thus, the feature descriptor is vulnerable to the precise location of the key pose.
			\item There are no constraints on the location of the segments with respect to the vehicle, which influences the way RANSAC estimates a registration. In forests and orchards the simple regularity can cause the step to struggle such as in frame 803 of Exp. C (\figref{fig:nsm_vs_icp_map_cornbury}), where the error is noticeably higher than the RMSE. This is still considered as a true localization (blue dot $\approx5\,m$ RMSE and $30^{\circ}$ rotation error), as the segments are matched correctly but the precise alignment was incorrect due to all the segments being located on one side of the vehicle. This will be addressed in our future work.
		\end{enumerate}
		
		\section{Conclusion and Future work}
		\label{sec:conclusion}
		In this paper, we presented Natural Segmentation and Matching, a novel approach for feature description and matching for global 
		localization in natural and urban environments. Our method exploits rigid objects which 
		are repeatable and distinctive in unstructured scenes which allows us to 
		successfully recognize previously visited places. We implemented our approach 
		and evaluated it on an urban, park, and heavily vegetated datasets. The experiments demonstrated that NSM performs comparably to state-of-the-art approaches in an urban environment, while also outperforming these algorithms in natural environments without specific parameter tuning. 
		
		In future plans we aim to extend the algorithm to learn environment specific features and to perform multi-seasonal evaluation. 

		\section{Acknowledgement}
		We would like to thank the authors of SegMatch~\cite{dub2017icra} for open sourcing their implementation and sharing valuable insights about it.
		
		\bibliographystyle{unsrt}

		\bibliography{library}
		
	\end{document}